\pdfoutput=1

\documentclass[11pt,a4paper]{article}
\usepackage{resources/template_2022/acl}
\usepackage{times}
\usepackage{latexsym}

\usepackage{microtype}

\usepackage{graphicx}
\graphicspath{ {./} }
\usepackage{url}
\usepackage{xspace}
\usepackage{comment}
\usepackage{multirow}
\usepackage{booktabs} 
\usepackage{subfig}
\usepackage{soul}
\newcommand\rurl[1]{%
  \href{https://#1}{\nolinkurl{#1}}%
}

\usepackage[T1]{fontenc}

\usepackage[utf8]{inputenc}

\usepackage{microtype}

\title{Multi-Domain Targeted Sentiment Analysis}
\author{Orith Toledo-Ronen, Matan Orbach, Yoav Katz, Noam Slonim \\
IBM Research \\
\texttt{\{oritht, matano, katz, noams\}@il.ibm.com}\\ 
}

\newcommand{\percentage}[1]{$#1$\%\xspace}

\newcommand{\sentenceQuote}[1]{\emph{"#1"}}

\newcommand{\paren}[1]{\left({#1}\right)}

\newcommand{\TSAtaskName}[0]{TSA\xspace}

\newcommand{\TargetExtraction}[0]{TE\xspace}
\newcommand{\TargetSentiment}[0]{SC\xspace}

\newcommand{\fOne}[0]{$F_1$\xspace}

\newcommand{\labelName}[1]{\textup{#1}\xspace}

\newcommand{\positiveLabel}[0]{\labelName{P}}
\newcommand{\negativeLabel}[0]{\labelName{N}}

\newcommand{\noneLabel}[0]{\labelName{O}}

\newcommand{\YasoName}[0]{\textsc{YASO}\xspace}
\newcommand{\CCName}[0]{\textsc{MD}\xspace}
\newcommand{\AmazonName}[0]{\textsc{Amazon}\xspace}
\newcommand{\YelpName}[0]{\textsc{Yelp}\xspace}

\newcommand{\OpinosisName}[0]{\textsc{Opinosis}\xspace}
\newcommand{\SemEvalName}[0]{\textsc{SemEval}\xspace}

\newcommand{\semEvalOneFourName}[0]{\textsc{SE}\xspace}

\newcommand{\MamsName}[0]{\textsc{MAMS}\xspace}
\newcommand{\domainName}[1]{\textit{#1}\xspace}

\newcommand{\baselineName}[1]{{#1}\xspace}

\newcommand{\baselineNameBERTB}[0]{\baselineName{BERT-B}}
\newcommand{\baselineNameBERTMLM}[0]{\baselineName{BERT-MLM}}
\newcommand{\baselineNameSENTIX}[0]{\baselineName{SENTIX}}
\newcommand{\baselineNameBERTPT}[0]{\baselineName{BERT-PT}}

\newcommand{\LabeledDataModelName}[0]{\textsc{LD}\xspace}
\newcommand{\WeakLabelsModelName}[0]{\textsc{WL}\xspace}
\newcommand{\MLMtaskName}[0]{MLM\xspace}

\newcommand{\tableRef}[1]{Table~\ref{#1}}
\newcommand{\sectionRef}[1]{\S\ref{#1}}
\newcommand{\figureRef}[1]{Figure~\ref{#1}}

\definecolor{darkgreen}{rgb}{0, 0.5, 0}

\newcommand{\targetTermExample}[1]{\textbf{#1}}

\newcommand{\forFinal}[1]{}

\begin{document}
\maketitle
\begin{abstract}

Targeted  Sentiment Analysis (TSA) is a central task for generating insights from consumer reviews. Such content is extremely diverse, with sites like Amazon or Yelp containing reviews on products and businesses from many different domains.
A real-world \TSAtaskName system should gracefully handle that
diversity. This can be achieved by a multi-domain model -- one that is robust to the domain of the analyzed texts, and performs well on various domains. 
To address this scenario, we present a multi-domain \TSAtaskName system based on augmenting a given training set with diverse weak labels from assorted domains.
These are obtained through self-training on the \YelpName reviews corpus.
Extensive experiments with our approach on three evaluation datasets across different domains demonstrate the effectiveness of our solution. 
We further analyze how restrictions imposed on the available labeled data affect the performance, 
and compare the proposed method to the costly alternative of manually gathering  diverse TSA labeled data. Our results and analysis show that our approach is a promising step towards a practical domain-robust TSA system.

\end{abstract}
\section{Introduction}
\label{sec:introduction}

Customer reviews of products and businesses provide insights for both consumers and companies. They help companies understand customer satisfaction or guide marketing campaigns, and aid consumers in their decision-making.
Sentiment analysis plays a central role in the analysis of such material,
by aiming to understand the sentiment expressed in a review document or in  
a single review sentence \citep{liu2012sentiment}.
Beyond these high-level trends, identifying the  sentiment towards a specific product feature or an entity is important. 
Such a fine-grained analysis includes the key task of Targeted Sentiment Analysis (\TSAtaskName), aimed at detecting sentiment-bearing terms in texts and classifying the sentiment towards them.
For example, in the sentence \sentenceQuote{The room was noisy, but the food was tasty,} the targets are \textbf{room} and \textbf{food} with negative and positive sentiments, respectively.
Our focus in this work is on \TSAtaskName of user reviews in English.

A real-world \TSAtaskName system has to successfully process diverse data.
From toothbrushes to phones, airline companies to local retailers, the online content today covers a broad range of reviews in many domains.
Ideally, a system for such a \emph{multi-domain} scenario should be able to cope with inputs from any domain, those that were seen during training,  
and, perhaps more importantly, those that were not. 

To the best of our knowledge, this work is the first to pursue \TSAtaskName in a multi-domain setup, intending to support input from multiple unknown domains.
Many previous works have used the \emph{in-domain} setup of training and testing on data from the same domain (e.g. \citet{li-etal-2019-exploiting}).  
Newer works focus on the \emph{cross-domain} setup, yet most have explored a \emph{pairwise} evaluation of training on one source domain and evaluating on a single known target domain (e.g. \citet{rietzler-etal-2020-adapt,gong-etal-2020-unified}).

Broadly, multi-domain learning \citep{emnlp/joshi12/multi-domain} includes training and evaluation using data from multiple domains (e.g. \citet{EMNLP/dredze2008/multi-domain, acl20/qin20/multi-domain-dialog-systems, scl/dai2021/multidomain-dialogs}).
Sometimes, it is assumed that the input texts are accompanied by a domain label (e.g. \citet{emnlp/joshi12/multi-domain}).
Here, we do not assume a domain label is given -- this has the advantage of allowing easier practical use of our model, without having  to specify the domain as part of the input.
In other cases, evaluation is limited to domains represented in training, or otherwise performed in a zero-shot setup only on unseen domains \citep{acl/wang20/multi-domain-NER}. 
Our system handles both cases simultaneously, processing data from domains well-represented in the training data as well as from unseen domains.

For practical reasons, implementing a multi-domain system with a single model 
that can handle all domains is desirable. This can save valuable resources such as memory or GPUs, which are in high demand by contemporary language models (LMs).
For example, it is impractical to expect that an online service providing \TSAtaskName analysis will have a per-domain model, each keeping its many parameters in memory, along with perhaps a set of pre-allocated GPUs. 
Our goal is therefore to have a single multi-domain model that performs well on both seen and unseen domains. 
This is reminiscent of works in multilingual NLP that develop a single model that handles multiple languages (e.g. M-BERT released by
\citet{devlin2019bert}, \citet{liang2020xglue}, \citet{toledo-ronen-etal-2020-multilingual}).

A possible approach to our setting is training on a diverse \TSAtaskName dataset, potentially encompassing many of the domains that the system is applied to. However, obtaining such a dataset is a challenge.
The existing \TSAtaskName datasets are limited in their diversity, and the collection of a new large scale diverse \TSAtaskName dataset is complex \citep{orbach2021yaso}.

The road we take is therefore based on augmenting a \TSAtaskName dataset of limited diversity with assortment of weak labels, through self-training 
-- one of the earliest ideas for utilizing unlabeled data in training \citep{chapelle2009semi}. 
To show that our approach is feasible, we performed an extensive empirical evaluation with several LMs that were fine-tuned with labeled data from the \SemEvalName dataset of \citet{pontiki-2014-semeval} (henceforth \semEvalOneFourName). This dataset is limited to two domains: restaurants or laptops.
Each model went through several self-training iterations and evaluated
on three \TSAtaskName publicly available
datasets: \semEvalOneFourName, the \MamsName dataset of restaurant reviews 
\cite{jiang-etal-2019-challenge}, and the \YasoName dataset of open-domain reviews \citep{orbach2021yaso}.

As part of our evaluation, we created two new \TSAtaskName resources.
The first is an annotation layer on top of the \YasoName dataset, specifying the domain of each review. This allows a per-domain evaluation providing insights on the performance of seen and unseen domains. The second resource is a set of manually annotated  \TSAtaskName reviews, which can be an ad-hoc diverse \TSAtaskName training set, an alternative to the proposed method.
We show that even in the presence of such data in training our approach is valuable. Both resources are available online.\footnote{\rurl{github.com/IBM/yaso-tsa}}

In summary, the main contributions of this work are:
(i) the first exploration of \TSAtaskName in a multi-domain setup; 
(ii) demonstrating the feasibility of multi-domain \TSAtaskName 
by an extensive evaluation on three datasets and the use of self-training;
(iii) the release of additional \TSAtaskName resources: a new annotation layer for the \YasoName dataset, and a set of fully annotated reviews.

\section{Related work}
\label{sec:related_work}

\vspace{-3pt}
\paragraph{\TSAtaskName}
The \TSAtaskName task has been extensively studied in different scenarios. 
Some works considered it as a pipeline of two subtasks:
(i) aspect-term  extraction (\TargetExtraction) for identifying target terms in texts (e.g. \citet{li-etal-2018-aspect, xu-etal-2018-double}), and 
(ii) aspect-term sentiment classification (\TargetSentiment) for determining the sentiment towards a given target term (e.g. \citet{naacl/dai21/sentiment-classification/syntax-matter, li2019aspectSentimentClassification, wang-2018-aspectSentimentClassification}). 
Full \TSAtaskName systems may combine these building blocks 
by running \TargetExtraction and then \TargetSentiment in a pipeline. 
Others, like our system, use a single engine that provides an end-to-end solution to the whole task, and may be based on pre-trained language models (e.g.  \citet{li-etal-2019-exploiting, phan-ogunbona-2020-modelling}) or a generative approach \citep{yan-2021-generative-absa, acl/zhang21/towards-generative-absa}.
In a cross-domain setup, \TSAtaskName research includes \citet{acl/chen-qian-21/-bridge} on \TargetExtraction, \citet{rietzler-etal-2020-adapt} on \TargetSentiment, \citet{CL/wang2020/cross-domain-target-and-opinion-extraction, coling/pereg2020/cross-domain-target-and-opinion-extraction} for joint \TargetExtraction and opinion term extraction and \citet{gong-etal-2020-unified} for the full \TSAtaskName task.
In contrast with our setup, these works all evaluate on one known domain.

\vspace{-3pt}
\paragraph{Domain Adaptation}
A plethora of domain adaptation (DA) methods have been 
developed for handling data from domains that are under-represented in training.
Several DA variants exist, of which the most common one handles a single known target domain.
For sentiment analysis, DA is especially important, as sentiment baring words tend to differ between domains \cite{ruder2017data}.
One promising DA approach is adjusting a given 
LM to a target domain using pre-training tasks performed on 
unlabeled data from that domain \cite{xu-etal-2019-bert, rietzler-etal-2020-adapt, zhou-etal-2020-sentix}. 
Another recently proposed direction of DA explored self-training for 
sentiment analysis (e.g. \citet{liu2021cycle}).

\vspace{-3pt}
\paragraph{Self Training} At the core of our approach is the iterative process of
self-training.  
This methodology has been successfully applied for varied research problems, e.g. object detection \cite{rosenberg_2005.107}, parsing \cite{mcclosky-etal-2006-effective}, handwritten digit recognition \cite{Lee2013PseudoLabelT} and image classification \cite{Zou2019ConfidenceRS} (see also the survey by \citet{triguero-2015-self}). 
Since the emergence of pre-trained LMs, several works have explored fine-tuning these models through self-training. Some examples are works on sentiment and topic classification \cite{yu-etal-2021-fine}, negation detection \cite{su-etal-2021-university}, toxic span detection \cite{suman-jain-2021-astartwice}, text classification \cite{karamanolakis-etal-2021-self} and machine translation \cite{sun-etal-2021-self}.

\section{Method}
\label{sec:method}
Our self-training approach augments a given \TSAtaskName training set with weak-labels (WL) generated from a large multi-domain corpus.
The process (depicted in \figureRef{fig:wl_genration}) starts by training an initial \TSAtaskName model on that given training data.
Then, that model produces \TSAtaskName predictions on a large  unlabeled corpus of diverse reviews. 
Finally, some of the predictions are selected and added as weak labels to the original training set. 
A new model is then trained with the augmented data, applied to produce new predictions on the unlabeled data, and 
the whole process (detailed below) can repeat for several iterations. 

\subsection{\TSAtaskName Engine}
We consider \TSAtaskName as a sequence tagging problem, where the  
model predicts a discrete label for each token of the input sequence.
The possible labels are: 
positive (\positiveLabel), negative (\negativeLabel) or none (\noneLabel). 
The first two labels represent tokens that are part of a sentiment
target, and the \noneLabel label represents all other non-target tokens. 
For example, given 
\sentenceQuote{Here is a nice electric car}, the desired output is  
the target \targetTermExample{electric car},  
identified from the output word-level sequence 
$\paren{\noneLabel, \noneLabel, \noneLabel, \noneLabel, \positiveLabel, \positiveLabel}$. 
During inference, for each sub-word piece within the input text, the labels scores outputted by the transformer model are converted into probabilities by applying softmax, and the highest probability label is selected.
The sub-word pieces predictions within each word are then merged by inducing the label of the first word piece with sentiment on the other word-pieces. Finally, consecutive word sequences having the same label (\positiveLabel or \negativeLabel) constitute one predicted target.

Our tagging scheme falls under the category of a unified tagging scheme \citep{li2019unified} with IO labels.
Previous works with a unified scheme used the more complex IOBES labels \citep{li2019unified, li-etal-2019-exploiting},
where the B and E labels designate the beginning and end of a target, respectively, and S represents a single token target.
Observing that the labeled data rarely includes two adjacent targets, the B and E labels were omitted (following  \citet{ijca/breck2007/identifying-opinion-expressions}).
The S label was excluded since in practice tokenization was to sub-word pieces, making the prediction of a single S label redundant.

\subsection{Unlabeled Data Set}
\label{subsec:data_set}
We use the \YelpName reviews data to create the weakly-labeled dataset for training. We start the process by extracting $2M$ sentences from the \YelpName corpus\footnote{\rurl{yelp.com/dataset}}. The corpus contains the text of the review documents and a list of business categories that correspond to each review. 
The reviews were initially selected at random, and then some reviews 
were removed by two conditions: 
reviews that are rated as \emph{not useful} (with useful=0) and
reviews of businesses with no business categories.
For each review, we assigned a single representative domain based on its 
business categories.  
The domain was determined by the first match between the review's categories and a predefined list of domains constructed from the categories in the corpus ordered by their popularity. 

Following the document-level filtering, each review was split into sentences, and the sentences were further filtered by:
1) length: only sentences with 10-50 words were selected; and
2) sentiment: at least one sentiment word should appear in the sentence. 
For the sentiment filter, we used a general-purpose lexicon  
-- 
the Opinion Lexicon \cite{hu_liu_sentiment_lex} that was automatically expanded by an SVM classifier and filtered 
as described in \citet{bar-haim-etal-2017-improving}. From that lexicon, we took all the sentiment words with score $S$ with confidence threshold of $|S|>0.7$, resulting with $7497$ sentiment words.

Finally, the representative domain of each review was assigned to all its selected sentences. 
Overall, we identified 18 different domains in the $2M$ extracted sentences, as shown in Table \ref{tab:yelp_data_stats}. 
We can see that $~60\%$ of the extracted data is from restaurants reviews, but 
the other $40\%$ of the data cover a variety of other domains. 
\begin{table}[t]
    \centering
    \small
    \begingroup
    \renewcommand{\arraystretch}{0.9}
    \begin{tabular}{rl|rl}
    \toprule
    \bf{Domain} & \bf{Sentences} & \bf{Domain} & \bf{Sentences} \\
    \midrule
    Restaurants & 1,195,156 &  Entertainment & 47,618  \\
    Food & 109,278 & Bars & 31,449  \\
    Beauty\&Spas & 106,023 & Pets & 26,679 \\
    Services & 102,471 & Local Flavor & 10,688 \\
    Travel & 92,600 & Education & 6,561  \\ 
    Shopping & 87,224 & Nightlife & 3,855  \\
    Automotive & 66,107 & Television & 2,170 \\
    Health & 60,768 & Religious & 1,468 \\ 
    Active Life & 49,094 & Media & 791 \\
    \bottomrule
    \end{tabular}
    \endgroup
    \caption{Data extracted from the \YelpName corpus with total of 2M sentences in 18 domains.}
    \label{tab:yelp_data_stats}
\end{table}

\subsection{Generating Weak labels}
\begin{figure*}[t]
    \centering
     \includegraphics[scale=0.54, trim={2.8cm 7.0cm 1cm 5.9cm},clip]{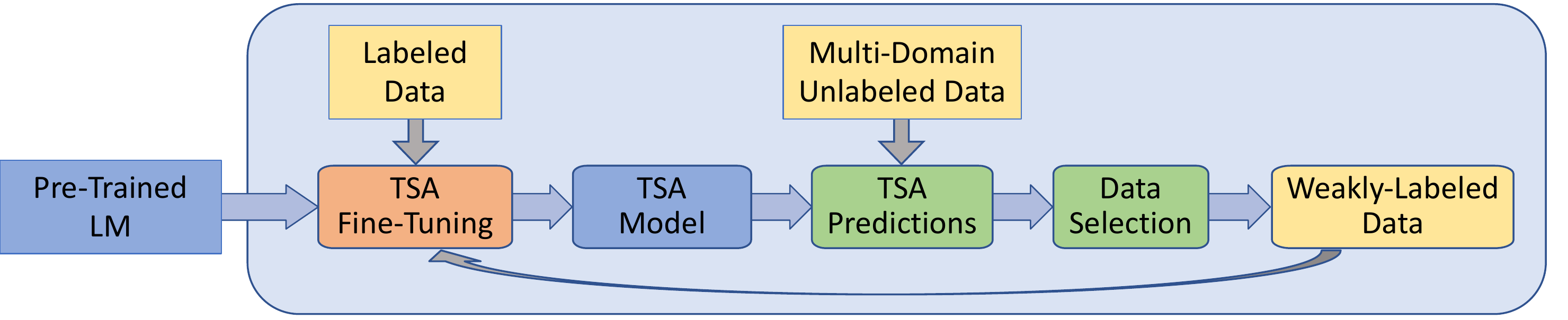}
    \caption{Weak labels generation and \TSAtaskName modeling process.}
    \label{fig:wl_genration}
\end{figure*}

The process, depicted in Figure \ref{fig:wl_genration},
starts by training a model on  
\TSAtaskName labeled data (henceforth, the LD model),  
followed by iteratively generating \TSAtaskName weak labels by self-training.
The initial \LabeledDataModelName model is used
for predicting \TSAtaskName target spans and sentiments on the  
unsupervised data. Each prediction is associated with a score $S$. 
We use it as a confidence score and select a subset of the sentences 
according to the following recipe: 
1) \textbf{targets}: 
sentences with targets that have confidence $S>0.9$
are selected if all other targets in the same sentence have 
$S<=0.5$. The high-confidence targets are 
added to the \TSAtaskName weak labels and the other 
predictions are ignored;   
2) \textbf{non-targets}: 
sentences with no predictions or if all the predicted targets have score $S<=0.5$ are selected and all the predictions are ignored.
To limit the amount of this part of the data, 
these sentences are randomly selected from $10\%$ of the data.
3) \textbf{domain balancing}:
the number of sentences per domain is limited to 
20k for each part of the data -- for sentences with 
targets and for those with no identified targets. 
This creates a balance between the representation of different domains in the data. Without balancing, about half of the 
selected data is from restaurants and other domains are under-represented. 

The selected sentences (those with \TSAtaskName weak labels and those with no targets) are then added to the labeled data, and a new \TSAtaskName model is fine-tuned and used for \TSAtaskName prediction and sentence selection over the 
entire unsupervised dataset in the next iteration. We repeated the  
process of \WeakLabelsModelName generation and model training 3 times  
and used the model from the third iteration for evaluation. 
The total number of sentences in the \WeakLabelsModelName data generated from the 2M sentences extracted from \YelpName 
is about 280k. 
This number depends on the initial \LabeledDataModelName model and on the number of iterations performed.

\section{Empirical Evaluation}
\label{sec:exp_results}

\subsection{Evaluation Data}
\paragraph{\YasoName}  
In \citet{orbach2021yaso}, we presented the \YasoName \TSAtaskName dataset comprising of user reviews from multiple sources.
This dataset covers reviews from many domains, and is thus a good choice for multi-domain evaluation.
While \YasoName allows an assessment on diverse reviews, its data is unbalanced between domains, thus biasing a standard evaluation 
towards the more common domains.
A per-domain evaluation is therefore complementary, and can help validate that a model performs well on all domains, not just the common ones.
Such an evaluation can also aid in discerning between performance on domains that are well-represented in the labeled data and ones that are unseen, thus verifying that the evaluated model performs well in both cases.

To facilitate such a per-domain evaluation, we augmented \YasoName with a domain label for each of its annotated reviews.
The assigned labels were produced automatically, when possible, or otherwise they were manually set by one of the authors. 
Since \YasoName contains annotated reviews from multiple sources, the assigned label depended on the source: reviews taken from the Stanford Sentiment Treebank \citep{socher-etal-2013-recursive, pang2005seeing} were assigned the \domainName{movies} domain label. 
Reviews from the \OpinosisName source \citep{ganesan2010opinosis} were assigned a label of \domainName{electronics}, \domainName{automotive} or \domainName{hotels}, based on the topic provided in that corpus for each review.
For example, reviews on \textit{transmission\_toyota\_camry\_2007} 
were assigned to \domainName{automotive}. 
In the \YelpName source, each review is associated with a list of business categories.
These categories were used as domain labels: we manually selected $8$ prominent categories as domains, and automatically matched the reviews to the domains using the category lists.
Reviews matched to multiple categories were manually examined and assigned the most fitting domain from the matched categories. 
Texts from the \AmazonName source \citep{keung2020multilingualAmazon} were manually read and labeled. 

Finally, the assigned domain labels were categorized into:
\domainName{restaurants} (with $400$ sentences), \domainName{electronics} ($412$), \domainName{hotels} ($161$), \domainName{automotive} ($144$), \domainName{movies} ($500$) and \domainName{other} ($596$).
This extra annotation layer of the \YasoName evaluation data 
is available online (see \sectionRef{sec:introduction}).
As suggested in \citet{orbach2021yaso},  \YasoName is used solely for evaluation.

\paragraph{\MamsName} \citet{jiang-etal-2019-challenge} collected the \MamsName dataset over restaurant reviews. 
In \MamsName, each  sentence has at least two targets\footnote{Called aspect terms in \citet{jiang-etal-2019-challenge}.} annotated with different sentiments. The sentiments are either \labelName{positive}, \labelName{negative} or \labelName{neutral}. 
To match our setup, the \labelName{neutral} labels were removed from these data.
The $500$ sentences of the \MamsName test set serve as an additional evaluation set. 

\paragraph{\semEvalOneFourName} \citet{pontiki-2014-semeval} created the popular \semEvalOneFourName dataset of restaurants and laptops reviews.  
We follow the standard split of \semEvalOneFourName into two sets with
$6072$ training sentences and $1600$ test sentences.
In each set, the sentences are balanced between the two domains.
As in \MamsName, the \labelName{neutral} labels were removed, as well as the \labelName{mixed} sentiment labels.

\subsection{Language Models}
The following four pre-trained LMs were used in our experiments:

\vspace{-3pt}
\paragraph{\baselineNameBERTB} \cite{devlin2019bert} The BERT-base uncased model with 110M parameters. 

\vspace{-3pt}
\paragraph{\baselineNameBERTMLM} 
To adjust \baselineNameBERTB to user reviews and sentiment analysis, 
we further pre-train it on the Masked Language Model (MLM) task, using the $2M$ review sentences extracted from  \YelpName (see \sectionRef{subsec:data_set}).
Our masking includes two randomly selected sets: 
(i) $15\%$ of the words in each sentence, as in \baselineNameBERTB; 
(ii) $30\%$ of the sentiment words in each sentence.
The sentiment words are taken from the union of two sentiment lexicons, one of \citet{bar-haim-etal-2017-improving} (with a confidence threshold of $0.7$), and the other created by \citet{toledo-ronen-etal-2018-learning} (with a confidence threshold of $0.5$, yielding $445$ words not present in the first lexicon).
Our masking of sentiment words is similar to the method used by \citet{zhou-etal-2020-sentix}, yet we do not use the emoticon masking.

\vspace{-3pt}
\paragraph{\baselineNameBERTPT} \cite{xu-etal-2019-bert} 
A variant of \mbox{\baselineNameBERTB} post-trained on the MLM and Next Sentence Prediction tasks using \YelpName data from the restaurants domain, and question answering data.\footnote{\rurl{huggingface.co/activebus/BERT-PT_rest}}

\vspace{-3pt}
\paragraph{\baselineNameSENTIX} \cite{zhou-etal-2020-sentix} 
A sentiment-aware language model for cross-domain sentiment analysis. 
This model was pre-trained with reviews from Yelp and Amazon, using an \MLMtaskName task that randomly masks sentiment words, emoticons, and regular words. 

\subsection{Experimental setting}

\paragraph{Training} Our fine-tuning
used a cross-entropy loss, 
the Adam optimizer \cite{adam-optimizer} with a learning rate of 3e-5 
and epsilon of 1e-8. The training process
was running with batch size of 32 on 2 GPUs with a maximum of 15 epochs and early stopping with min\_delta of $0.005$.
In each experiment, \percentage{20} of the training set sentences were randomly sampled and used as a development set.
The optimized metric on this set was the overall token \fOne classification rate.

\paragraph{Evaluation}
For each experiment, we trained $10$ models with different random seeds.
Then, the per-domain performance metrics
were computed for each run, and averaged for a final per-domain result (mean and standard deviation).
These per-domain results were macro-averaged to obtain the overall performance on each dataset.
As evaluation metrics we report
the precision (P), recall (R), and \fOne (mean and std), 
of exact match predictions.

\subsection{In-Domain Results}
\newcommand{\SemEvalInDomainName}[0]{\textbf{SE$_{R/L}$}}
\begin{table*}[ht]
    \small
    \centering
    \resizebox{\textwidth}{!}{ 
    \begingroup
    \renewcommand{\arraystretch}{0.8}
    \begin{tabular}{llcccccc}
\toprule
& & \multicolumn{3}{c}{\bf{Restaurants}} & \multicolumn{3}{c}{\bf{Laptops}} \\
\cmidrule(rl){3-5}
\cmidrule(rl){6-8}
\textbf{\textit{LM}} & \textbf{Train Set} & \textbf{P} & \textbf{R} & \textbf{F1} & \textbf{P} & \textbf{R} & \textbf{F1} \\
\midrule
\multirow{2}{*}{\textbf{\textit{\baselineNameBERTB}}} & \textbf{\SemEvalInDomainName}	& 67.7	&	77.3	&	72.1	$\pm$	0.8	&	55.9	&	65.8	&	60.4	$\pm$	1.3	\\
& \textbf{\SemEvalInDomainName+WL} & 74.0	&	75.3	&	\textbf{74.6}	$\pm$	0.5	&	63.7	&	63.4	&	\textbf{63.6}	$\pm$	0.8	\\
\midrule
\multirow{2}{*}{\textbf{\textit{\baselineNameBERTMLM}}} & 
\textbf{\SemEvalInDomainName} & 70.9	&	81.7	&	75.9	$\pm$	0.7	&	57.4	&	64.8	&	60.8	$\pm$	1.3	\\
& \textbf{\SemEvalInDomainName+WL} & 76.0	&	79.8	&	\textbf{77.8}	$\pm$	0.3	&	62.4	&	63.3	&	\textbf{62.8}	$\pm$	0.8	\\
\midrule
\multirow{2}{*}{\textbf{\textit{\baselineNameBERTPT}}} & 
\textbf{\SemEvalInDomainName} & 71.6	&	81.4	&	76.1	$\pm$	0.8	&	58.1	&	67.2	&	62.3	$\pm$	1.5	\\
& \textbf{\SemEvalInDomainName+WL} & 78.4	&	77.1	&	\textbf{77.7}	$\pm$	0.6	&	63.6	&	65.0	&	\textbf{64.2}	$\pm$	0.9	\\
\midrule
\multirow{2}{*}{\textbf{\textit{\baselineNameSENTIX}}} & 
\textbf{\SemEvalInDomainName} & 70.4	&	80.2	&	74.9	$\pm$	0.7	&	60.3	&	70.6	&	65.0	$\pm$	1.1	\\
& \textbf{\SemEvalInDomainName+WL} & 76.3	&	78.4	&	\textbf{77.4}	$\pm$	0.3	&	65.7	&	67.5	&	\textbf{66.6}	$\pm$	0.7	\\
\bottomrule
    \end{tabular}
    \endgroup
    }
    \caption{In-domain results on SE comparing fine-tuning of four \textbf{\textit{LM}}s with in-domain labeled data (\textbf{\SemEvalInDomainName}) and with self-training (\textbf{\SemEvalInDomainName+WL}). 
    }
    \label{tab:indomain_se14_results}
\end{table*}

Before showing the multi-domain results that are the focus of this work, we present the in-domain performance of our system on the widely-used \semEvalOneFourName evaluation data. These results, summarized in \tableRef{tab:indomain_se14_results}, serve as a sanity check for our system on a well-known benchmark in a well-explored setup. 

Explicitly, several single-domain 
models were created by fine-tuning each pre-trained 
LM with training data from one \semEvalOneFourName domain, 
either restaurants (R) or laptops (L).
These models, denoted SE$_{R/L}$, were evaluated 
on test data from the same domain they were trained on.
For \baselineNameBERTB, the results of this evaluation (top row of \tableRef{tab:indomain_se14_results}) were inline with previous works (cf. \citet{WANG2021178}).

For each LM, \tableRef{tab:indomain_se14_results} further shows 
the results with added WL data (SE$_{R/L}$+WL), created using the corresponding SE$_{R/L}$ model on the diverse \YelpName corpus. 
Interestingly, in all cases augmenting the training set with these WL improves results over the models trained without such data.

\subsection{Multi-Domain Results}
\begin{table*}[ht]
    \centering
    \resizebox{\textwidth}{!}{ 
    \begingroup
    \renewcommand{\arraystretch}{0.8}
    \begin{tabular}{llccccccccc}
\toprule
& & \multicolumn{3}{c}{\bf{\YasoName}} & \multicolumn{3}{c}{\bf{\MamsName}} & \multicolumn{3}{c}{\bf{\semEvalOneFourName}} \\
\cmidrule(rl){3-5}
\cmidrule(rl){6-8}
\cmidrule(rl){9-11}
\textbf{\textit{LM}} & \textbf{Train Set} & \textbf{P} & \textbf{R} & \textbf{F1} & \textbf{P} & \textbf{R} & \textbf{F1} & \textbf{P} & \textbf{R} & \textbf{F1}\\
\midrule
\multirow{2}{*}{\textbf{\textit{\baselineNameBERTB}}} & \textbf{SE}	& 	59.1	&	43.9	&	48.7	$\pm$	2.1	&	38.5	&	66.4	&	48.7	$\pm$	1.6	&	63.6	&	72.4	&	67.7	$\pm$	1.0	\\
& \textbf{SE+WL} &	68.5	&	45.9	&	\textbf{53.7}	$\pm$	1.1	&	46.5	&	62.7	&	\textbf{53.4}	$\pm$	0.7	&	67.6	&	71.7	&	\textbf{69.6}	$\pm$	0.7	\\
\midrule
\multirow{2}{*}{\textbf{\textit{\baselineNameBERTMLM}}} & \textbf{SE} &	60.5	&	46.0	&	50.6	$\pm$	1.5	&	38.4	&	69.2	&	49.3	$\pm$	1.2	&	65.1	&	73.7	&	69.1	$\pm$	0.8	\\
& \textbf{SE+WL} &	65.6	&	47.3	&	\textbf{54.0}	$\pm$	0.9	&	45.8	&	62.1	&	\textbf{52.7}	$\pm$	0.6	&	69.6 &	74.4	&	\textbf{71.9}	$\pm$	0.7	\\
\midrule
\multirow{2}{*}{\textbf{\textit{\baselineNameBERTPT}}} & \textbf{SE} & 	61.4	&	46.0	&	51.3	$\pm$	1.5	&	39.6	&	68.3	&	50.1	$\pm$	1.1	&	65.5	&	73.0	&	69.0	$\pm$	1.0	\\
& \textbf{SE+WL} &	68.6	&	48.1	&	\textbf{55.4}	$\pm$	1.0	&	45.2	&	61.5	&	\textbf{52.1}	$\pm$	0.7	&	69.6	&	74.3	&	\textbf{71.9}	$\pm$	0.7	\\
\midrule
\multirow{2}{*}{\textbf{\textit{\baselineNameSENTIX}}} & \textbf{SE} & 62.4	&	47.0	&	51.5	$\pm$	1.4	&	38.2	&	69.0	&	49.2	$\pm$	1.0	&	64.8	&	75.4	&	69.7	$\pm$	0.9	\\
& \textbf{SE+WL} &	69.8	&	44.9	&	\textbf{52.4}	$\pm$	0.7	&	44.7	&	61.1	&	\textbf{51.6}	$\pm$	0.3	&	71.5	&	74.9	&	\textbf{73.1}	$\pm$	0.5	\\
\bottomrule
    \end{tabular}
    \endgroup
    }
    \caption{Multi-domain results comparing the fine-tuning of four \textbf{\textit{LM}}s with labeled data only (\textbf{SE}) and with self-training (\textbf{SE+WL}), on the three evaluation datasets.
    }
    \label{tab:main_results}
\end{table*}

For the main evaluation of our approach, 
we fine-tuned each LM with the full \semEvalOneFourName training set 
(with data of both the R and L domains), generated the \WeakLabelsModelName data by self-training starting from the baseline model (\semEvalOneFourName), and then fine-tuned the final model (\semEvalOneFourName$+$\WeakLabelsModelName).
Table \ref{tab:main_results} presents the results obtained with these fine-tuned models, on \YasoName, \MamsName, and \semEvalOneFourName. 
In all cases, \fOne is improved by employing self-training.
For example, with \baselineNameBERTB, there is a $10\%$ relative gain in \fOne 
on \YasoName and \MamsName, and a $3\%$ relative gain on \semEvalOneFourName. 
Even with stronger base models such as \baselineNameSENTIX or \baselineNameBERTPT that incorporate domain knowledge into the language model, we see gains of several points in \fOne by adding the \WeakLabelsModelName data. 
The gain in \fOne is mostly due to gain in precision, sometimes at some cost  
in recall (specifically for \MamsName).
The variance of \fOne across the different training runs is significantly reduced. 

\begin{table*}[ht]
    \centering
    \resizebox{\textwidth}{!}{ 
    \begingroup
    \renewcommand{\arraystretch}{0.8}
    \begin{tabular}{llccccccccc}
\toprule
& & \multicolumn{3}{c}{\bf{\YasoName}} & \multicolumn{3}{c}{\bf{\MamsName}} & \multicolumn{3}{c}{\bf{\semEvalOneFourName}} \\
\cmidrule(rl){3-5}
\cmidrule(rl){6-8}
\cmidrule(rl){9-11}
\textbf{\textit{System}} & \textbf{Train Set} & \textbf{P} & \textbf{R} & \textbf{F1} & \textbf{P} & \textbf{R} & \textbf{F1} & \textbf{P} & \textbf{R} & \textbf{F1}\\
\midrule
\textbf{\textit{Gong-BASE}} & \textbf{SE}	&	66.3	&	43.8 &	50.8	$\pm$	1.1	&	42.5 &	68.4 	& 52.4 	$\pm$	0.7	&	69.6	&	74.4	&	\textbf{71.9}	$\pm$	0.7	\\
\textbf{\textit{Gong-UDA}} & \textbf{SE $\rightarrow$ YELP}	&	60.8	&	48.5	&	52.7	$\pm$	1.1	&	38.6	&	72.5	&	50.4	$\pm$	0.2	&	65.1	&	77.4	&	70.7	$\pm$	1.4	\\
\midrule
\multirow{2}{*}{\textbf{\textit{Ours}}} & \textbf{SE}	&	59.1	&	43.9	&	48.7	$\pm$	2.1	&	38.5	&	66.4	&	48.7	$\pm$	1.6	&	63.6	&	72.4	&	67.7	$\pm$	1.0	\\
& \textbf{SE+WL}	&	68.5	&	45.9	&	\textbf{53.7}	$\pm$	1.1	&	46.5	&	62.7	&	\textbf{53.4}	$\pm$	0.7	&	67.6	&	71.7	&	69.6	$\pm$	0.7	\\
\bottomrule
    \end{tabular}
    \endgroup
    }
    \caption{Multi-domain results with \citet{gong-etal-2020-unified} (baseline (\textbf{\textit{BASE}}) and the \textbf{\textit{UDA}} approach; average of 3 training runs) compared with our results (baseline (\textbf{SE}) and self-training (\textbf{SE+WL})). All the results are with \textbf{\textit{\baselineNameBERTB}}.}
    \label{tab:gong_results}
\end{table*}

\begin{figure*}[t]
    \centering
    \includegraphics[trim={1.5cm 0cm 1.5cm 0cm}]{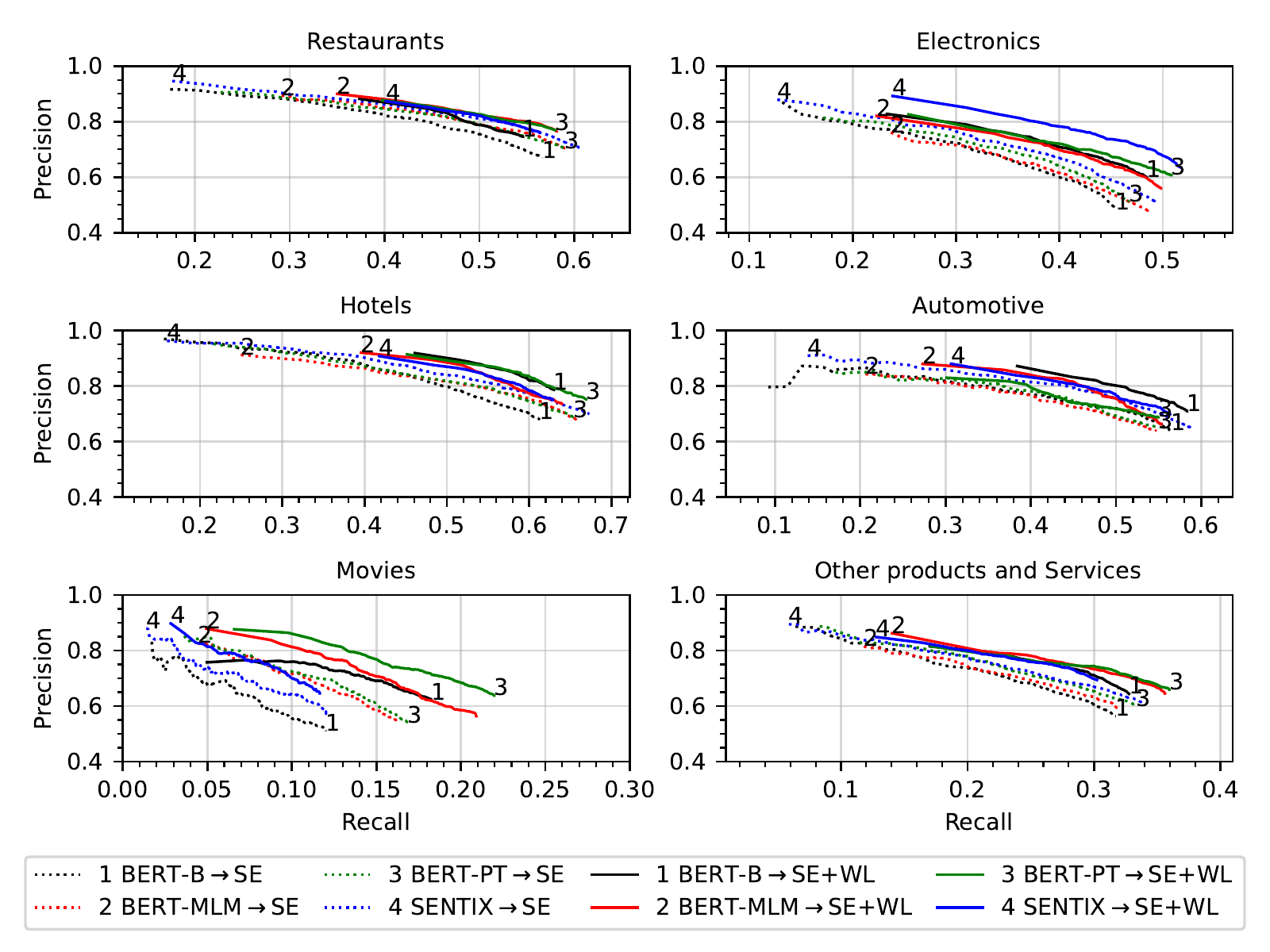}
    \caption{Per-domain precision-recall curves on \YasoName 
    of fine-tuning the four LMs (numbered 1-4 at the end of each line) with self-training (solid lines, tuned on SE+WL data) and without it (dotted lines, tuned on SE data).  
    }
    \label{fig:main_results_yaso}
\end{figure*}

\begin{figure}[t]
    \centering
    \includegraphics[width=\linewidth,trim={0.4cm 1.28cm 0.35cm 0.8cm},clip]{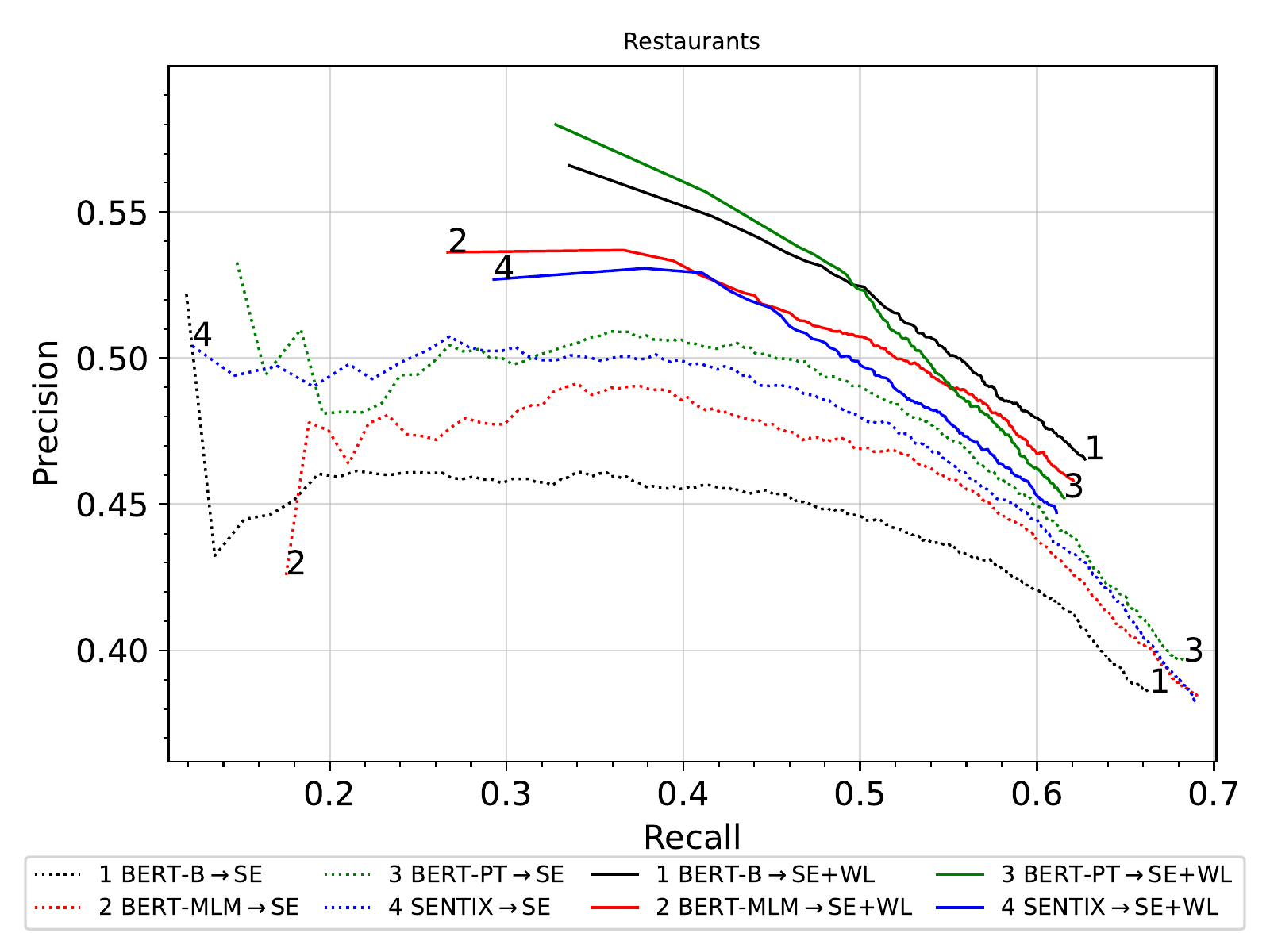}
    \caption{Precision-recall results on \MamsName 
    with self-training (solid lines) and without it (dotted lines).
    The graph uses the same legend as Figure \ref{fig:main_results_yaso}.
    }
    \label{fig:main_results_mams}
\end{figure}

\figureRef{fig:main_results_yaso} further details per-domain results on \YasoName, showing precision/recall curves for each fine-tuned LM with and without self-training.
As above, each curve is the average of 10 per-run curves.
In most cases, 
the self-trained models outperform the initial corresponding fine-tuned 
\semEvalOneFourName models.
This result is also apparent in \figureRef{fig:main_results_mams} for \MamsName. Here, although recall is decreased for self-trained models their precision 
is significantly improved across the entire curve.

Next, we compare our self-supervision approach with the cross-domain \TSAtaskName work of \citet{gong-etal-2020-unified}.\footnote{\rurl{github.com/NUSTM/BERT-UDA}}
To adjust their system to a multi-domain setup, we use the full \semEvalOneFourName training set (R and L) as the labeled data from the source domain (as in our system), and a random sample from the \YelpName unlabeled data to represent the target domain. 
The number of sentences in the sample equals the size of the training set, as in their experiments. The sample was also balanced across all $18$ domains.

\tableRef{tab:gong_results} includes the results of this comparison.
On \YasoName, their baseline results (Gong-BASE) improve when integrating their domain adaptation components (Gong-UDA), yet they are lower than with our
self-supervision results (except for on \semEvalOneFourName).

\subsection{Impact of the Initial \LabeledDataModelName Model}
The quality and quantity of the \TSAtaskName labeled data used for training the initial \TSAtaskName 
model are important factors for
the quality of the weak labels induced by its predictions. 
This, in turn, affects the quality of the entire self-training process.
This experiment explores this effect, by imposing restrictions on the training set of the initial \TSAtaskName model.

In this context, we experimented with three variants.
One model was fine-tuned with half of the \semEvalOneFourName data (SE$_h$), selected at random from each domain, such that overall the samples
were balanced between the two domains.
Two more models were fine-tuned with \semEvalOneFourName data from one domain -- restaurants (SE$_R$) or laptops (SE$_L$).
For all models, the number of sentences in the training set was half the size of the full \semEvalOneFourName data.

\tableRef{tab:ld_content_results} summarizes the results of our experiments with these models, focusing on the \baselineNameBERTMLM pre-trained model.
As expected, training on a single domain, or with half of the data, leads to lower performance.
The results on the \MamsName restaurants data are typical for a cross-domain setup. When training on laptop reviews alone, recall drops almost entirely to $2.6$, and self-training improves upon that poor performance to some extent. 
Overall, across all datasets and all training data starting points, performance consistently improves when self-supervision is used.

\newcommand{\seRestaurants}[0]{\textbf{SE$_R$}}
\newcommand{\seLaptops}[0]{\textbf{SE$_L$}}
\newcommand{\seHalf}[0]{\textbf{SE$_h$}}
\begin{table*}[ht]
    \centering
    \resizebox{\textwidth}{!}{ 
    \begingroup
    \renewcommand{\arraystretch}{0.8}
    \begin{tabular}{llccccccccc}
\toprule
& & \multicolumn{3}{c}{\bf{\YasoName}} & \multicolumn{3}{c}{\bf{\MamsName}} & \multicolumn{3}{c}{\bf{\semEvalOneFourName}} \\
\cmidrule(rl){3-5}
\cmidrule(rl){6-8}
\cmidrule(rl){9-11}
\textbf{\textit{LM}} & \textbf{Train Set} & \textbf{P} & \textbf{R} & \textbf{F1} & \textbf{P} & \textbf{R} & \textbf{F1} & \textbf{P} & \textbf{R} & \textbf{F1}\\
\midrule
\textbf{\textit{\multirow{8}{*}{\baselineNameBERTMLM}}} & \textbf{SE}	&	60.5	&	46.0	&	50.6	$\pm$	1.5	&	38.4	&	69.2	&	49.3	$\pm$	1.2	&	65.1	&	73.7	&	69.1	$\pm$	0.8	\\
& \textbf{SE+WL}	&	65.6	&	47.3	&	\textbf{54.0}	$\pm$	0.9	&	45.8	&	62.1	&	\textbf{52.7}	$\pm$	0.6	&	69.6	&	74.4	&	\textbf{71.9}	$\pm$	0.7	\\
\cmidrule(rl){2-11}
 & \seHalf	&	58.3	&	46.0	&	49.9	$\pm$	1.3	&	35.6	&	67.7	&	46.7	$\pm$	1.0	&	62.5	&	72.4	&	67.1	$\pm$	1.0	\\
& \textbf{\seHalf+WL}	&	66.0	&	42.4	&	\textbf{50.3}	$\pm$	1.2	&	44.2	&	61.0	&	\textbf{51.3}	$\pm$	0.7	&	68.7	&	67.4	&	\textbf{68.0}	$\pm$	0.8	\\
\cmidrule(rl){2-11}
 & \seRestaurants	&	61.8	&	40.5	&	47.1	$\pm$	2.1	&	36.5	&	68.7	&	47.6	$\pm$	1.4	&	58.1	&	58.8	&	57.9	$\pm$	1.3	\\
& \textbf{\seRestaurants+WL}	&	69.1	&	40.2	&	\textbf{49.4}	$\pm$	1.1	&	42.8	&	64.9	&	\textbf{51.6}	$\pm$	0.7	&	64.2	&	59.3	&	\textbf{61.2}	$\pm$	0.9	\\
\cmidrule(rl){2-11}
 & \seLaptops	&	60.4	&	21.7	&	27.9	$\pm$	2.5	&	37.3	&	2.6	&	4.9	$\pm$	2.1	&	70.0	&	36.5	&	37.9	$\pm$	2.2	\\
& \textbf{\seLaptops+WL}	&	67.0	&	22.5	&	\textbf{30.6}	$\pm$	1.5	&	51.2	&	4.4	&	\textbf{8.1}	$\pm$	1.0	&	71.5	&	36.8	&	\textbf{40.5}	$\pm$	1.0	\\	
\bottomrule
    \end{tabular}
    \endgroup
    }
    \caption{Multi-domain results comparing the fine-tuning of \textbf{\textit{\baselineNameBERTMLM}} with labeled data only (\textbf{SE}) and with self-training (\textbf{SE+WL}), 
    with four initial models trained with data from: the entire SE data (\textbf{\semEvalOneFourName}), 
    half the data from each of the SE domains (\seHalf), or a single SE domain -- restaurants (\seRestaurants) or laptops (\seLaptops).
    }
    \label{tab:ld_content_results}
\end{table*}

\subsection{Diversifying the Training Set}
An alternative to our weak-labeling approach is diversifying the \TSAtaskName training set by manual labeling.
To explore this option, we collected an ad-hoc \TSAtaskName training dataset 
that contains $952$ sentences of reviews from multiple domains.
The collection started with reviews written by crowd annotators in a given domain, on a topic of their choice.\footnote{We refrain from the annotation of existing proprietary data due to the legal restrictions imposed on its redistribution with additional annotation layers.} 
The reviews were then annotated for \TSAtaskName by 
asking annotators to mark all sentiment-bearing targets in each sentence. 
This step is similar to the candidates annotation phase described in \citet{orbach2021yaso}. However, unlike in our previous work,
the detected candidates we collected were not passed through 
another verification step, to reduce costs. This results in noisier data, unfit for evaluation purposes, yet a manual examination has shown it is of sufficient quality for training. 

\tableRef{tab:cc_results} shows the performance obtained 
using this new dataset for training. The collected multi-domain 
labels (henceforth \CCName) were combined with the \semEvalOneFourName data for fine-tuning the \baselineNameBERTB and \baselineNameBERTMLM models. 
Comparing the results of fine-tuning with data from limited domains (\semEvalOneFourName) to fine-tuning with the additional \CCName data,
performance significantly improves on the diverse \YasoName evaluation set.
On \MamsName the improvement is small, presumably because the restaurants domain is well covered in the \semEvalOneFourName training set.
On the \semEvalOneFourName test set the improvement is negligible or non-existent.
When comparing our approach using the \WeakLabelsModelName data 
to the \CCName alternative, there is an improvement in F1 
on both \MamsName and \semEvalOneFourName, yet results on \YasoName are somewhat lower. 
However, the precision achieved by our approach is consistently better on all three evaluation sets compared to the alternative method.  
Similar trends are observed using \baselineNameBERTMLM. 
Overall, the results with \WeakLabelsModelName are better or close to those with \CCName, with the advantage that no manual labeling is required. 

\begin{table*}[ht]
    \centering
    \resizebox{\textwidth}{!}{ 
    \begingroup
    \renewcommand{\arraystretch}{0.8}
    \begin{tabular}{llccccccccc}
\toprule
& & \multicolumn{3}{c}{\bf{\YasoName}} & \multicolumn{3}{c}{\bf{\MamsName}} & \multicolumn{3}{c}{\bf{\semEvalOneFourName}}\\
\cmidrule(rl){3-5}
\cmidrule(rl){6-8}
\cmidrule(rl){9-11}
\textit{\textbf{LM}} & \textbf{Train Set} & \textbf{P} & \textbf{R} & \textbf{F1} & \textbf{P} & \textbf{R} & \textbf{F1} & \textbf{P} & \textbf{R} & \textbf{F1}\\
\midrule
\multirow{2}{*}{\textbf{\textit{\baselineNameBERTB}}} 
  & \textbf{SE+MD}	&	62.2	&	50.2	&	\textbf{54.2}	$\pm$	1.8	&	38.8	&	67.6	&	49.3	$\pm$	1.0	&	62.5	&	73.7	&	67.6	$\pm$	1.0	\\
 & \textbf{SE+WL}	&	68.5	&	45.9	&	53.7	$\pm$	1.1	&	46.5	&	62.7	&	\textbf{53.4}	$\pm$	0.7	&	67.6	&	71.7	&	\textbf{69.6}	$\pm$	0.7	\\
\midrule
\multirow{2}{*}{\textbf{\textit{\baselineNameBERTMLM}}} 
 & \textbf{SE+MD}	&	63.1	&	51.1	&	\textbf{55.1}	$\pm$	1.3	&	39.3	&	67.2	&	49.6	$\pm$	0.8	&	64.9	&	74.5	&	69.4	$\pm$	1.0	\\
  & \textbf{SE+WL}	&	65.6	&	47.3	&	54.0	$\pm$	0.9	&	45.8	&	62.1	&	\textbf{52.7}	$\pm$	0.6	&	69.6	&	74.4	&	\textbf{71.9}	$\pm$	0.7	\\
\bottomrule
    \end{tabular}
    \endgroup
    }
    \caption{A comparison of fine-tuning two \textbf{\textit{LM}}s with data augmented through self-training (\textbf{SE+WL}) or combined with a multi-domain \TSAtaskName dataset (\textbf{SE+\CCName}).
    }
    \label{tab:cc_results}
\end{table*}

\section{Manual Error Analysis}
\label{sec:data_analysis}

The automatic evaluation reported above is based on exact-span matches, and may be too strict in some cases.
For example, in \sentenceQuote{The best thing about this place is the different sauces,} the \YasoName labeled data contains the target \targetTermExample{the different sauces}, thus counting a prediction of \targetTermExample{sauces} as an error. 
Alternative evaluation options may circumvent this problem.  
For
example, the above prediction would be considered as correct using overlapping span matches. However, changing the automatic evaluation can introduce new issues and and may be too lenient.
Continuing the above example, with an overlapping span match, a prediction of the entire sentence is also considered as correct.

Due to these issues, we complement the automatic evaluation with a manual one, comparing the output of an initial LD model to its self-trained counterpart.
The error analysis was performed on one experimental setup,  
with the \baselineNameBERTMLM pre-trained model and the entire \semEvalOneFourName dataset for training the \LabeledDataModelName model.
We further focus on the \YasoName dataset: for each model, $30$ predictions considered as errors by the automatic evaluation were randomly sampled from each of the $6$ domains.
One of the authors categorized these predictions into one of four options: 
invalid target, correct target identified with wrong sentiment or span, borderline target that can be accepted, and a clearly correct target.
The latter are presumably due to the strictness of the exact-matches based evaluation.

\tableRef{tab:error_analysis} presents the results of this manual analysis.
Overall, the self-trained model (SE+WL) predicts less non-targets.
Moreover, it identifies more valid targets than the baseline model. As for the other two categories of errors, the borderline 
and wrong span/sentiment, the two models are on par. 
These results emphasize the importance of manual error analysis, and show that even in this detailed analysis, which goes beyond the labeling information
available in the \YasoName evaluation set, we find that the multi-domain model with the \WeakLabelsModelName is better.

\begin{table}[]
    \centering
    \begingroup
    \renewcommand{\arraystretch}{0.8}
    \begin{tabular}{lcc}
\toprule
\textbf{Error Analysis} & \textbf{SE} & \textbf{SE+WL} \\
\midrule
Invalid target	     & 27.2\% &	17.8\% \\
Wrong sentiment/span & 18.3\% & 18.3\% \\
\midrule
Borderline target  & 14.4\% &	16.1\% \\ 
Correct target     & 40.0\% &	47.8\% \\  
\bottomrule
    \end{tabular}
    \endgroup
    \caption{Error analysis results on randomly selected wrong predictions on \YasoName evaluation. Predictions are obtained by the \MLMtaskName baseline model fine-tuned with the SE data (\emph{left}) and with SE+WL data (\emph{right}).}
    \label{tab:error_analysis}
\end{table}

\section{Conclusion} 

This work addressed a multi-domain \TSAtaskName setting in which a system is trained on data from a small number of domains, 
and is applied to texts from any domain. 
Our proposed method has employed self-learning to augment an existing \TSAtaskName dataset with weak labels obtained from a large corpus.

An empirical evaluation of our approach has demonstrated that the self-supervision technique, often used when having a training set of limited size, is also effective for enhancing the diversity of the training data.
Specifically, our results show that the self-trained multi-domain model
consistently improves performance, for various underlying LMs,
and with different starting points: 
data from two domains, removing half of the data, or restricting 
to only one domain.
Interestingly, even in the presence of a diverse \TSAtaskName labeled data, our approach was comparable to the performance obtained with that data.
This allows avoiding the burden and costs associated with manual \TSAtaskName data collection. 

In addition to finding targets and their sentiments, other
related tasks aim to extract the corresponding opinion term 
\citep{DBLP:conf/aaai/Peng20/target-sentiment-opinion-triplet-task},  identify the relevant aspect category \citep{wan2020target}, or both \cite{cai-2021-acos-new-task}.
As future work, our 
approach may be applied to these more complex tasks as well.
Similarly, it may be useful for developing a multilingual \TSAtaskName system, by utilizing weak labels produced on unlabeled reviews data in non-English languages.

\section*{Acknowledgments}
We wish to thank Artem Spector for the development of the experimental infrastructure. We also thank the anonymous reviewers for their insightful comments and feedback.

\bibliography{main}
\bibliographystyle{resources/template_2022/acl_natbib}

\end{document}